\documentclass{article}

\usepackage{PRIMEarxiv}

\usepackage[utf8]{inputenc} 
\usepackage[T1]{fontenc}    
\usepackage{hyperref}       
\usepackage{url}            
\usepackage{booktabs}       
\usepackage{amsfonts}       
\usepackage{nicefrac}       
\usepackage{microtype}      
\usepackage{lipsum}
\usepackage{fancyhdr}       
\usepackage{graphicx}       
\graphicspath{{media/}}     
\usepackage{comment}

\usepackage{enumitem}
\usepackage{bm}
\usepackage{amssymb}
\usepackage{caption}
\usepackage{subcaption}
\usepackage{amsmath}

\title{Activity Classification Using Unsupervised Domain Transfer from Body Worn Sensors}

\author{
  Chaitra Hedge \\
  School of Electrical and Computer Engineering \\
  Georgia Institute of Technology \\
  Atlanta, GA USA\\
  \texttt{chegde@gatech.edu} \\
   \And
  Gezheng Wen\\
  \\
  Amazon Services Inc \\
  Seattle, WA USA \\
  \texttt{wengz@amazon.com} \\
  \And
  Layne C. Price
  \thanks{\textit{Corresponding Author}: 
  Layne C. Price, \texttt{prilayne@amazon.com}}\\
  \\
  Amazon Services Inc \\
  Seattle, WA USA \\
  \texttt{prilayne@amazon.com} \\
}

\begin{document}
\maketitle

\begin{abstract}
Activity classification has become a vital feature of wearable health tracking devices. As innovation in this field grows, wearable devices worn on different parts of the body are emerging. To perform activity classification on a new body location, labeled data corresponding to the new locations are generally required, but this is expensive to acquire. In this work, we present an innovative method to leverage an existing activity classifier, trained on Inertial Measurement Unit (IMU) data from a reference body location (the source domain), in order to perform activity classification on a new body location (the target domain) in an unsupervised way, i.e. without the need for classification labels at the new location. Specifically, given an IMU embedding model trained to perform activity classification at the source domain, we train an embedding model to perform activity classification at the target domain by replicating the embeddings at the source domain. This is achieved using simultaneous IMU measurements at the source and target domains. The replicated embeddings at the target domain are used by a classification model that has previously been trained on the source domain to perform activity classification at the target domain. We have evaluated the proposed methods on three activity classification datasets PAMAP2, MHealth, and Opportunity, yielding high F1 scores of 67.19\%, 70.40\% and 68.34\%, respectively when the source domain is the wrist and the target domain is the torso.
\end{abstract}


\section{Introduction}
Body-worn measurement devices have become ubiquitous in both consumer and health contexts, as they can provide real-time monitoring and inference on a person's activities, as well as their general health and wellness.  Human activity recognition has been shown to be valuable in various fields, such as healthcare \cite{9393204} \cite{9055403}, patient monitoring \cite{gjoreski2016accurately} \cite{6063364} \cite{8861371} \cite{syed2022deep}, gait analysis \cite{8248364} \cite{8434292}, surveillance and safety applications \cite{taha2015human} \cite{gowsikhaa2012suspicious} \cite{al2017novel}, and gesture recognition \cite{nunez2018convolutional} \cite{kim2019imu}. Most of these human activity recognition methods use machine learning or statistical methods based on accelerometers and/or inertial measurement unit (IMU) sensors present onboard the device, either independently or in conjunction with other sensors, such as photoplethysmogram \cite{electronics10141715} \cite{9910287}.\\

\begin{figure}[t!]
    \includegraphics[width=0.7\textwidth]{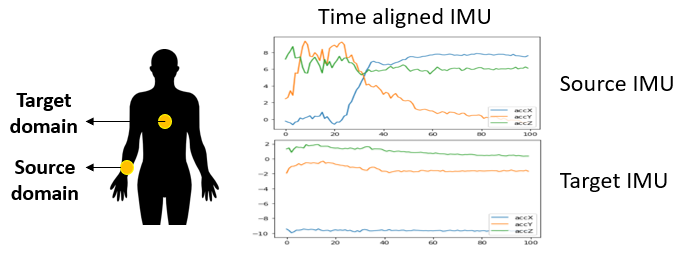}
    \centering
    \caption{An example of source domain (e.g., wrist) and target domain (e.g., torso) body locations. This work uses a model that is trained to perform activity classification on the source domain to do the same on the target domain. It is assumed that a well performing activity classifier $M_S$ has been trained on the source domain data and that a small amount of unlabeled data is available for the target domain with simultaneous readings from the source domain. $M_S$ is retrained using this unlabeled data to obtain $M_T$ which is used for activity classification on the target domain. } 
    \label{fig:overview}
\end{figure}

Wrist-worn measurements devices such as smart watches and fashionable accessories can easily host an IMU, and are hence the most common consumer-grade activity measurement tool. There are plenty of good quality labeled data from the wrist, on which numerous advanced activity classification algorithms have been trained. There is growing interest in body-worn measurement devices that are not placed on the wrist, e.g., devices placed on the ankle, torso, or waist \cite{storm2016gait} \cite{dobkin2011reliability}, but the amount of labeled data from these body locations are scarce. In this paper we propose a new method to use an existing activity classifier that has been trained on IMU data obtained in a source domain (e.g. the wrist) to perform activity classification predictions on a target domain (e.g. the torso) in an unsupervised way. Fig.~\ref{fig:overview} displays example traces of unlabeled, simultaneously collected IMU data from the source and target domains.\\

Our method falls under the umbrella of unsupervised domain transfer methods. It does not require labeled data at the target body site, and only relies on simultaneous IMU measurements at the source and target body sites. A large amount of such data can be easily obtained in various day-to-day scenarios from a large number of subjects, thus capturing a wide variety of complex activities. Additionally, our method can be generalized to any source and target combinations. A unique characteristic of our method is that an existing activity classifier trained on the source body location can be applied at the target body location without making changes to the architecture of the classifier. Thus, advantages of the trained classifier can be retained, and applied directly to the target body site, which would significantly save cost and modeling effort. This is especially useful in commercial settings where well performing activity classification models already exist for wrist IMU data, and are efficient enough to run on edge devices.

\section{Related work}

Multiple measurement modalities have been used for activity recognition, including body worn sensors like IMUs \cite{HASSAN2018307, 8586957, informatics9030056}, accelerometers \cite{8368718}, and off-body sensors like cameras \cite{almaadeedHAR, gleasonUntrimmed}.  Many supervised machine learning methods in this field have shown promise, such as support vector machines \cite{HASSAN2018307, 8368718, KOPING2018248}, convolutional neural networks (CNNs) \cite{KHAN2021107671, informatics5020026, s19173731}, long-short term memory networks (LSTMs) \cite{SAINI201899}, attention mechanisms \cite{KHAN2021107671, ijcai2019p431}, and a combination of CNNs and recurrent neural networks \cite{informatics9030056, ijcai2019p431}.\\

While most supervised machine learning approaches for activity recognition require large amount of labeled data, collecting labeled data from human subjects is expensive and time-consuming. Many works in the literature have identified this shortcoming and have proposed methods to circumvent this problem to the extent possible.  Examples include active learning to reduce the amount of data required to be labeled \cite{s19030501}, semi-supervised feature extraction \cite{8586957}, and ensembles of models trained on small quantities of sensor data \cite{8767027}.\\

Sometimes, a large amount of labeled data are available for activity classification at a particular body location (the source domain) and a small amount of labeled data is available at the target body location where an activity classifier is desired. Transfer learning approaches have been commonly used in such cases \cite{transferWang2018, stratWang2018, chen2019cross}, and other scenarios, such as between different user populations \cite{Litransfer2022}, application domains and sensor modalities \cite{moralesTransfer2016}. Stratified transfer learning methods have been popular in such applications \cite{stratWang2018, chen2019cross}. Another approach that has been used is to add a new layer to an existing pre-trained model in addition to retraining the model \cite{transferWang2018}. \\ 

Unsupervised domain adaptation is an approach used when no labeled data is available at the target body site. Faridee et. al. \cite{faridee2022strangan} used a spatial transformer to learn affine transformations that transform the target IMU to have discriminative features similar to those for the source domain IMU. Chen et. al.
\cite{chen2019motiontransformer} proposed a more complex model with encoder, generator, decoder and predictor modules to converts raw IMU signals into domain invariant hidden representations. Mu et. al.
\cite {mu2020unsup} used domain adversarial neural networks (DANN) \cite{GaninDANN2016} and extended multi-source DANN \cite{zhao2018adversarial} \cite{PeiMSDANN2018} to perform unsupervised domain adaptation for gait event detection and abnormal gait pattern recognition, in which DANN projects IMU data from the source and target domains into a common subspace using a feature extractor, domain classifier and a task specific classifier. \\

\section{Unsupervised Domain Transfer Approach}
\label{sec:approach}

\begin{figure}[tb]
    \includegraphics[width=0.48\textwidth]{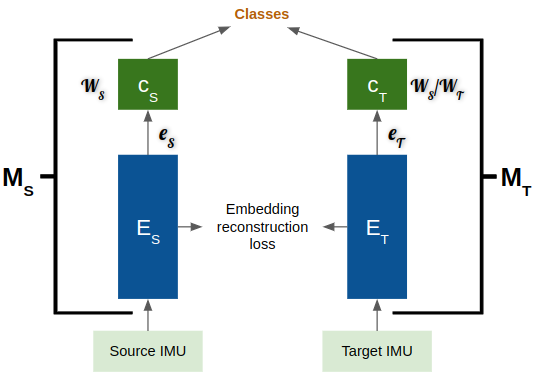}
    \centering
    \caption{Diagram showing the unsupervised domain adaptation approach. The source-domain model $M_S$ consists of the embedding extractor module $E_S$ and the classifier module $C_S$. Similarly the target-domain model model $M_T$ consists of the embedding extractor module $E_T$ and classifier module $C_T$. The embedding extractor modules $E_S$ and $E_T$ generate embeddings $e_S$ and $e_T$ from the input IMU signals. These embeddings are used by the classifier modules $C_S$ and $C_T$ to output the activity classes for the source and target domains respectively. An embedding reconstruction loss is used to train $E_T$ with IMU signals simultaneously recorded at the source and target body locations such that $e_T$ is similar to $e_S$. The weights $W_S$ of classifier $C_S$ can be applied to classifier $C_T$ when no target domain labels are available. In the case where target domain labels are available, $C_T$ can be fine-tuned using the labels to have weights $W_T$.}
    \label{fig:flowchart}
\end{figure}

The schematic of our proposed unsupervised domain transfer approach is shown in Fig.~\ref{fig:flowchart}.
Our approach consists of two models, the source-domain model $M_S$ and the target-domain model $M_T$. The source-domain model, $M_S$, can either be an existing pre-trained model or it can be trained in a supervised way using a source-domain dataset $D_S$ containing a collection of IMU or accelerometer signals $\alpha_{S}(t)$ and labels $l_{S}(t)$ and can be defined as:
\begin{equation}
    D_S = \{[\alpha_{S}(t), l_{S}(t)]\}
\end{equation}
$M_S$ can be decomposed into an embedding extractor module $E_S$ and a classification module $C_S$. The embedding extractor module $E_S$ generates embeddings $e_S$ from the input IMU signals while the classification module uses these embeddings to output the final classified labels. 
$M_S$ can be defined as:
\begin{equation}
    M_{S}^{\theta}(\alpha_{S}(t))=\hat{l_{S}}(t),
\end{equation}
where $\theta$ are the real parameters of this model and $\hat{l_S}$ are the model's estimated output labels.
$M_S$ should have the following structure:
\begin{equation}
    M_S^\theta(\alpha_S) = \phi(W_S \, e_S^\theta(\alpha_S)),
    \label{eqn:m_source}
\end{equation}
where $\phi$ is a multi-dimensional logistic function, $W_S$ is a real-valued weight matrix, and $e_S^\theta \in \mathbb{R}^d$ is the $d$-dimensional embedding for the signal $\alpha_S$ produced by the embedding extractor module ($E_S$) of the model.\\

\noindent Similarly, the target-domain model $M_T$ can be decomposed into an embedding extractor module $E_T$ and a classification module $C_T$. The embedding extractor module $E_T$ generates embeddings $e_T$ from the input IMU signals while the classification module uses these embeddings to output the final classified labels. 
$M_T$ can be defined as:
\begin{equation}
    M_T^\psi = \phi(W_T \, e_T^{\psi}(\alpha_T)),
    \label{eqn:m_target}
\end{equation}
where $W_T$ is a real-valued weight matrix, $e_T \in \mathbb{R}^d$, and $\psi$ are real-valued parameters of the model.\\

The goal of our procedures is to train $E_T$ such that the produced embeddings $e_T$ are as similar to $e_S$ as possible for each activity class. 
For this, we use a dataset $D_{ST}$ that has simultaneuos IMU or accelerometer signals  measured at the source body location and the target body location, but does not necessarily contain activity labels. These time series signals at the source and target body locations are $\alpha_S(t)$ and $\alpha_T(t)$,  respectively. This dataset $D_{ST}$ can be defined as:
\begin{equation}
    D_{ST} = \{[\alpha_{S}(t), \alpha_{T}(t)]\}.
\end{equation}
$E_S$ provides a set of features, $e_S^\theta(\alpha_s)$, that are relevant for predicting the labels $l_S$. In this work, we further make the critical assumption that one or more of the features in $e_S$ are relevant for making predictions for the target-domain labels $l_T$. In other words, we assume that most dimensions in the source domain embeddings $e_S$ can be reconstructed from the target domain signals $\alpha_T(t)$.  It is important to note that only certain activity classes would have this property. For example, if the source domain is the wrist and the target domain is the torso, activities such as walking and jumping contain discriminating information that is present in both domains, as both body sites physically participate in such activity. In contrast, activities like typing or clapping have relevant information only in the wrist but not in the torso; typing and clapping are uninformative in the target domain. Here, in the scope of this paper, we restrict our attention to activity labels that can be mutually predicted from both source and target domain measurements. Therefore, our approach is to obtain target-domain embeddings $e_T^\psi$ by
minimizing a loss function $L$ that has a global minimum where $e_T(\alpha_T) = e_S(\alpha_S)$ for $[\alpha_T, \alpha_S] \in D_{ST}$:

\begin{equation} \label{eq:minimize_loss}
    \hat{\psi} = \underset{\psi} {argmin} \,\, L[e_S^\theta(\alpha_S), e_T^\psi(\alpha_T) ]
\end{equation}
Choices of the loss function $L$ can be varied; e.g., in Section \ref{sec:exp} we consider mean-absolute (MAE), mean-squared (MSE), and mean-squared-logarithmic error (MSLE), as well as cosine similarity.  Regularizations such as $L_1$ and $L_2$ regularizations can be also applied to promote the reconstruction of subsets of the source-domain embeddings that are most relevant to making predictions in the target domain.\\

Finally, we can obtain an unsupervised activity classifier at the target domain that predicts the source domain labels $l_S$ by setting $W_T = W_S$ in Eq.~\eqref{eqn:m_target}.  Alternatively, if target-domain labels $l_{ST}$ are available in $D_{ST}$, $W_T$ can be obtained by a second supervised fine-tuning step with a classification-relevant loss function, similarly to how $M_S$ is trained.

\begin{figure}[!t]
    \includegraphics[width=0.75\textwidth]{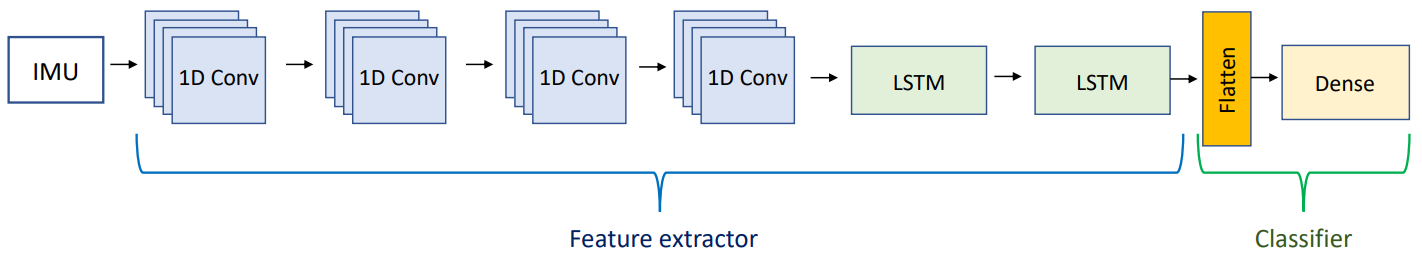}
    \centering
    \caption{Architecture diagram used for $M_S$ and $M_T$ obtained from \cite{ordonez2016deep}. The layers until the final LSTM layer are considered to make up the embedding extractor modules $E_S$ and $E_T$, while the rest of the layers comprise the classifier modules $C_S$ and $C_T$. The source and target domain models, $M_S$ and $M_T$, have the same architecture and hyperparameters.}
    \label{fig:m1}
\end{figure}

\section{Implementation}

The implementation of the approach discussed in Section \ref{sec:approach} is explained here.
The models $M_S$ and $M_T$ in Eqs~\eqref{eqn:m_source} and \eqref{eqn:m_target} are comprised of two components: the embedding extractor and the classifier.  For both models $M_S$ and $M_T$, we used the state-of-the-art convolution-LSTM architecture proposed by Ord{\'o}{\~n}ez \& Roggen \cite{ordonez2016deep}, as shown in Fig.~\ref{fig:m1}. The key advantages of the chosen architecture are its high performance and ability to disambiguate closely related activities while being a relatively light-weight model. The model consists of 4 convolutional layers with 64 feature maps and ReLU activation, 2 LSTM layers with 128 cells, and a dense layer with softmax activation as the last layer. The layers up to the final LSTM layer are considered to make up the embedding extractor module ($E_S$ and $E_T$), while the rest of the layers comprise the classifier module ($C_S$ and $C_T$). Note that the architecture and hyperparameter settings for $M_T$ and $M_S$ are the same. \\

For each of the publicly available datasets described later in Sect.~\ref{sect:data}, we followed the methods in Ord{\'o}{\~n}ez \& Roggen \cite{ordonez2016deep} and trained the corresponding reference source-domain model $M_S$ in a supervised manner using cross-entropy loss function and mini-batch gradient descent with RMSprop. Note that Ord{\'o}{\~n}ez \& Roggen \cite{ordonez2016deep} used data from multiple on-body sensors to train their model (i.e., all the sensors available in the datasets), while we only used data from the wrist sensors. For example, for the Opportunity dataset with IMU readings from 7 body locations, \cite{ordonez2016deep} used all 63 channels of data for training $M_S$ whereas we used only the 9 channels from the right wrist. 

The embeddings for the target-domain model $e_T$ are found by minimizing a reconstruction loss function, as shown in Eq.~\eqref{eq:minimize_loss}.  In Sect.~\ref{sec:exp} we compare results from various loss functions, including MSE, MAE, MSLE, and cosine similarity, along with $L_1$ and $L_2$ regularizations. We initialized the parameters of the target-domain model with those from the source-domain model, and used mini-batch gradient descent with RMPprop for optimization.

\section{Data}
\label{sect:data}
\begin{figure}[!t]
    \includegraphics[width=0.4\textwidth]{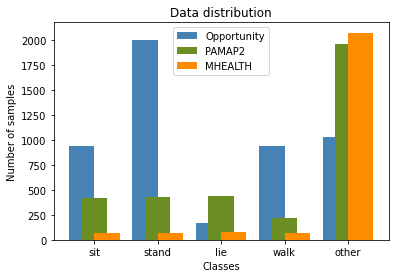}
    \centering
    \caption{Class distribution across the five overlapping labels (\emph{sit},  \emph{stand}, \emph{lie down}, \emph{walk} and \emph{other}) for Opportunity, PAMAP2 and MHEALTH datasets. Opportunity dataset has a higher number of samples than MHEALTH and PAMAP2 datasets. PAMAP2 and MHEALTH datasets have a much larger proportion of their samples in the  \emph{other} class as compared to the rest of the classes. Each dataset is imbalanced.}
    \label{fig:all_dist}
\end{figure}

We used three publicly available activity classification datasets for evaluating the proposed method: Opportunity \cite{roggen2010collecting}, PAMAP2 \cite{reiss2012introducing} and MHEALTH \cite{banos2014mhealthdroid}. Given the relatively higher quantity of data in the Opportunity dataset, we present our results on the Opportunity dataset in details, and use the PAMAP2 and MHEALTH datasets to further validate the observations. As all three datasets have measurements at the wrist and torso/chest, we chose to use the wrist as the source domain and the torso as the target domain for our analysis. 
The activity classes that overlap in all three datasets are \emph{sitting}, \emph{standing}, \emph{lying down}, \emph{walking} and \emph{other}. For a major part of our analysis, we used these five overlapping labels so as to maintain uniformity across the datasets. The class distribution of these five labels for each of the datasets are shown in Fig.~\ref{fig:all_dist}.
We describe each dataset below.

\begin{itemize}
    \item{The Opportunity dataset \cite{roggen2010collecting} were collected in a sensor-rich environment, with on-body sensors as well as external sensors that were placed on objects that study participants interacted with. A custom motion jacket was worn by each participant that was fitted with IMUs that measured seven body locations: right lower arm, left lower arm, right upper arm, left upper arm, back, right shoe, and left shoe. We used the IMU readings from the right lower arm and the back in our analysis. This dataset has four subjects that performed two sets of experiments: activities of daily living in a routine morning, and repeated scripted sessions following a protocol. Activity classifications labels are available at multiple granularity levels: locomotion labels, high-level activities, mid-level gestures, and low-level left/right hand actions. We used only the locomotion activities for our analysis, which consists of the following activities: sit, stand, walk, lie down, and other.}
    \item{The PAMAP2 dataset \cite{reiss2012introducing} consists of 9 subjects wearing 3 IMU sensors over the dominant arm's wrist, chest, and dominant leg’s ankle. Subjects in this dataset engaged in 18 day-to-day and sporting activities, including lying down, standing, walking, running, computer work, car driving, and others. For our analysis, we only focus on using the IMU readings from the dominant wrist and the chest.}
    \item{The MHEALTH dataset \cite{banos2014mhealthdroid} has 10 subjects wearing an accelerometer and ECG sensor on the chest, and IMUs on the left ankle and right lower arm. The subjects performed 12 sets of physical activities, which included standing still, sitting, lying down, walking, and others. For our analysis, we used the accelerometer sensors on the chest and right wrist. Note that we focused on using the accelerometer readings instead of IMU readings for this dataset since the chest was fitted with an accelerometer instead of an IMU.}
\end{itemize}

The sampling frequency of the Opportunity dataset was 30 Hz, and to achieve the same sampling frequency across the three datasets, we downsampled the PAMAP2 and MHEALTH dataset to 30Hz. Windows of data were then created such that each window had 100 samples, i.e., approximately 3 seconds of non-overlapping data. All datasets were made to have the same units for accelerometer ($m/s^2$), gyroscope ($deg/s$) and magnetometer ($\mu T$) readings. \\

For each dataset, 30\% of the dataset were randomly chosen for training the source-domain model $M_S$, where the wrist was considered to be the source-domain. An additional 50\% of the data were randomly chosen for the unsupervised training to obtain the target-domain model $M_T$ at the chest/torso, and the remaining 20\% was used for testing.

\section{Experiments \& Results}
\label{sec:exp}

\subsection{Performance metrics}
\label{subsec:exp1}

To assess the performance of the proposed method, we computed accuracy, precision, recall and F1 scores on the provided activity classes. The precision, recall and F1 score for each of the activity classes were calculated separately by considering the classification task as a one-vs-rest problem. The scores were then averaged across the classes to obtain the final precision, recall and F1 scores. For each of the three dataset, we evaluated the performance metrics with three experiments:
\begin{itemize}
\item{\emph{Supervised source model on the source domain.---}  The source domain model $M_S$ trained on the source-domain data $D_S$.  This is standard supervised training, which is used to gauge the overall performance potential of the model architecture on each of the three datasets.}
\item{\emph{Supervised source model on the target domain.---} The source domain model $M_S$ trained on the source-domain data $D_S$, but evaluated without any fine-tuning on the target-domain data $D_{ST}$. This assesses how well $M_S$ performs on $D_{ST}$, and provides a untrained baseline to compare the performance of trained $M_T$ with.}
\item{\emph{Unsupervised target model on the target domain.---} The target-domain model $M_T$ trained on the simultaneous source-domain and target-domain data $D_{ST}$ using our unsupervised domain transfer method in Sect.~\ref{sec:approach}.}
\end{itemize}

\begin{table*}[!t]
\centering
\caption{Activity classification performance metrics of 1) supervised source model on the source domain, i.e. $M_S$ on $D_S$, 2) supervised source model on the target domain, i.e. $M_S$ on $D_{ST}$ and 3) unsupervised target model on the target domain, i.e. $M_T$ on $D_{ST}$ for the three datasets. Training on $D_{ST}$ as part of the modeling of $M_T$ helps achieve better performance than $M_S$ applied to  $D_{ST}$, implying that the embeddings carry useful overlapping information between the source and target domains that the model $M_T$ is able to pick up on.}
\label{tab:5class_results}
\begin{tabular}
{|p{1.3cm}|p{1cm}|p{1cm}|p{1cm}|p{1cm}|p{1cm}|p{1cm}|p{1cm}|p{1cm}|p{1cm}|}
\hline
Metrics & \multicolumn{3}{|c|}{Opportunity} & \multicolumn{3}{|c|}{PAMAP2} & \multicolumn{3}{|c|}{MHEALTH} \\
\hline
 & $M_S$ on $D_S$ & $M_S$ on $D_{ST}$ & $M_T$ on $D_{ST}$ & $M_S$ on $D_S$ & $M_S$ on $D_{ST}$ & $M_T$ on $D_{ST}$ & $M_S$ on $D_S$ & $M_S$ on $D_{ST}$ & $M_T$ on $D_{ST}$\\
\hline
Accuracy & 76.13 & 24.01 & 65.13 & 94.47 & 61.00 & 77.26 & 92.64 & 80.16 & 91.20\\
\hline
Precision & 73.28 & 14.01 & 66.50 & 94.24 & 38.02 & 73.98 & 73.10 & 19.63 & 69.20\\
\hline
Recall & 79.60 & 26.2 & 67.89 & 92.85 & 41.29 & 67.15 & 88.33 & 24.96 & 67.50\\
\hline
F1 score & 76.31 & 18.26 & \textbf{67.19} & 93.54 & 39.59 & \textbf{70.40} & 80.00 & 21.98 & \textbf{68.34}\\
\hline
\end{tabular}
\end{table*}

Table \ref{tab:5class_results} shows the accuracy, precision, recall and F1 score for each of these three models for all the three datasets. The supervised training of $M_S$ on $D_S$ yields the highest scores for the PAMAP2 dataset. Lower performance on the MHEALTH dataset could be partially attributed to the smaller volume of training data and fewer channels of data being used (as stated in Section \ref{sect:data} we only used three accelerometer channels for MHEALTH dataset as opposed to nine IMU channels for the other two datasets). As demonstrated in \cite{10.1145/3090076}, activities in the Opportunity dataset were less structured and closer to real-life scenarios, hence is the most difficult dataset among the three, which resulted in the lowest performance of $M_S$ on $D_S$. The same performance trend across the three datasets is also observed in the other two experiments,  $M_S$ on $D_{ST}$ and $M_T$ on $D_{ST}$. \\

When comparing the model performance on the target domain, the target domain model $M_T$ trained on $D_{ST}$ performs significantly better in all the three datasets than the source domain model being directly applied to the target domain i.e., $M_S$ on $D_T$. This shows that the embedding extractor $E_{T}$, when trained on $D_{ST}$, is able to learn similar discriminative information as in the source domain embedding $e_S$ for constructing the target domain embeddings $e_T$. Thus, fine-tuning an existing activity classification model, as part of our unsupervised method, could be potentially sufficient for performing the classification task at a new body location. However, as the performance of $M_T$ on $D_{ST}$ across the three datasets are in the same order as of $M_S$ on $D_S$, the potential performance increase from such fine-tuning could be partially limited by the performance of the original supervised model from the source location.\\

\begin{figure*}[!t]
     \centering
     \begin{subfigure}[!t]{0.4\textwidth}
         \centering
         \includegraphics[width=\textwidth]{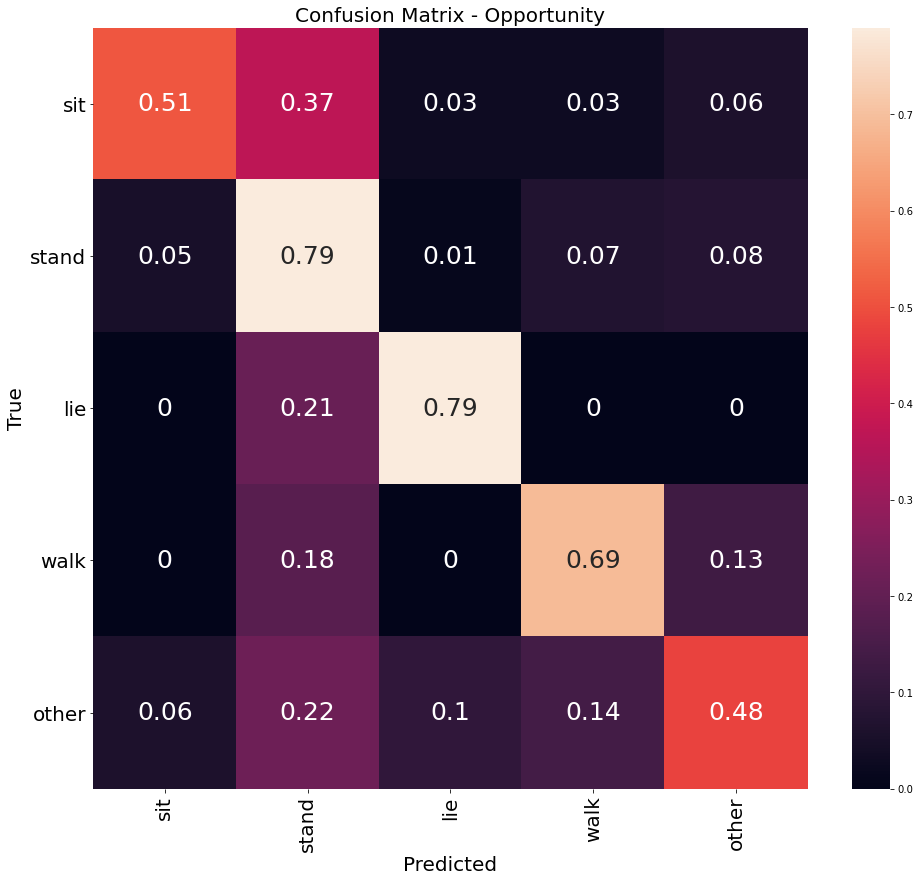}
         \caption{Confusion matrix}
         \label{fig:confusion_opp}
     \end{subfigure}
     \begin{subfigure}[!t]{0.5\textwidth}
         \centering
         \includegraphics[width=\textwidth]{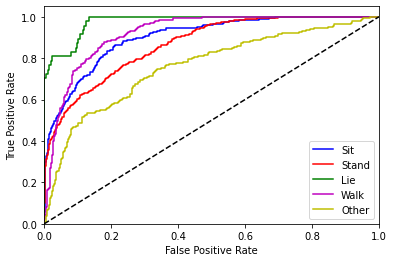}
         \caption{ROC curve} 
         \label{fig:roc_opp}
     \end{subfigure}
    \caption{(a) The confusion matrix of $M_T$ on the Opportunity dataset highlights the difference in classification performance for different classes. (b) ROC curves for each activity class in the Opportunity dataset. The classification of a class is more accurate if the overlapping information between $e_S$ and $e_T$ is more discriminative when compared to other classes.}
    \label{fig:confusion_roc_opp}
\end{figure*}

Fig. \ref{fig:confusion_opp} shows the confusion matrix from $M_T$ on the Opportunity dataset. It can be seen that the models performed better for some activities classes like \emph{stand} and \emph{lie down} than for the other classes. Many \emph{sit} samples are wrongly classified by the model as \emph{stand}. The \emph{other} class is commonly misclassified as \emph{sit}, \emph{stand}, \emph{lie} and \emph{walk}. Fig. \ref{fig:roc_opp} shows the corresponding ROC curves for each of the activity classes, where the model achieved the highest AUC for \emph{lie} while the lowest AUC for \emph{other}. Generally, $M_T$ would perform better on activities that are not very similar, and on activities where the common information between embeddings in the target and source domain is richer. 
The counterpart confusion matrices and ROC curves for the other two datasets can be found in the Appendix \ref{appendix:confusion} and \ref{appendix:roc}, respectively.

\subsection{Impact of quantity of training data}
\label{subsec:exp2}

To get a sense of how the size of unsupervised training data $D_{ST}$ may impact the performance of the target-domain model $M_T$, we trained $M_T$ separately with 15\%, 33\%, 66\% and 100\% of the samples from $D_{ST}$.

\begin{figure}[!t]
    \includegraphics[width=0.4\textwidth]{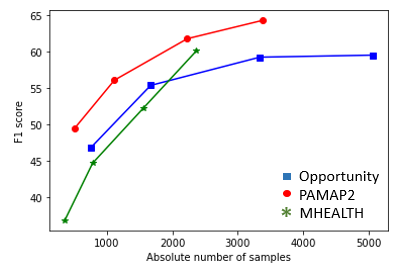}
    \centering
    \caption{F1 scores for different number of samples from each dataset. The green stars represent MHEALTH dataset, red kdots represent PAMAP2 dataset and blue squares represent Opportunity dataset. The trend seen is that the F1 score increases as the amount of training data increases for each dataset. Comparison across datasets is not possible due to differences in factors like sensor types, sensor placements, etc. } 
    \label{fig:absolute_samp}
\end{figure}

Fig. \ref{fig:absolute_samp} shows the F1 scores for each dataset against the number of randomly chosen samples used for this unsupervised training. 
The F1 score increases as the number of samples increases within each of the three datasets. As seen in the graph of the Opportunity dataset, such increase in performance seems to plateau beyond when the number of training samples exceeds a certain point. Here, the F1 performance stops improving substantially after approximately 3000 samples are included.  However, as briefly discussed in Section \ref{subsec:exp1}, the three datasets cannot be easily compared for performance due to key differences in the data collection process and protocol, sensors used, sensor placement, and thus the variation of difficulty in their respective activity classification. 

\subsection{Switching roles of source and target domains}
\label{subsec:exp3}

We repeated the experiment \ref{subsec:exp1} by switching the source and target domain body locations. That is, the torso was considered as the source body location and the wrist was considered as the target body location. This is to test our hypothesise that our model relies on the overlapping information between the source and target domains to perform classification at the target domain; hence, when the source and target domain are switched, the performance of the new classification model on the new target domain should be comparable, assuming the new $M_S$ has similar performance as the old one. Accordingly, $M_S$ was trained in the same supervised fashion on the torso IMU data and $M_T$ was trained using the wrist IMU data and corresponding torso embeddings. \\

Table \ref{tab:5labels_results_switched} shows the accuracy, precision, recall and F1 scores for this experiment \ref{subsec:exp3} for all three datasets. 
The performance of $M_S$ on $D_S$ remained similar for the Opportunity dataset when trained on the torso. However the performance for the PAMAP2 and MHEALTH datasets dropped. $M_S$ on $D_S$ performs as well as or worse when trained on the torso as compared to when trained on the wrist. One possible cause is movement data loss in the torso, such as arm swinging, for certain activities. The performances of $M_T$ follow the same performance trend as of $M_S$, achieving the highest performance with the PAMAP2, followed by the Opportunity and MHEALTH datasets.

\begin{table*}[!t]
\centering
\caption{Performance metrics as in Table~\ref{tab:5class_results} when the source and target domains were switched, i.e. the source domain is the torso and target domain is the wrist. There was a decrease in performance of $M_S$ on $D_S$ which led to a decrease in performance of $M_T$ on $D_{ST}$ when compared to Table \ref{tab:5class_results}.} 
\label{tab:5labels_results_switched}
\begin{tabular}
{|p{1.3cm}|p{1cm}|p{1cm}|p{1cm}|p{1cm}|p{1cm}|p{1cm}|p{1cm}|p{1cm}|p{1cm}|}
\hline
Metrics & \multicolumn{3}{|c|}{Opportunity} & \multicolumn{3}{|c|}{PAMAP2} & \multicolumn{3}{|c|}{MHEALTH} \\
\hline
  & $M_S$ on $D_S$ & $M_S$ on $D_{ST}$ & $M_T$ on $D_{ST}$ & $M_S$ on $D_S$ & $M_S$ on $D_{ST}$ & $M_T$ on $D_{ST}$ & $M_S$ on $D_S$ & $M_S$ on $D_{ST}$ & $M_T$ on $D_{ST}$\\
\hline
Accuracy & 74.41 & 20.13 & 63.98 & 83.00 & 51.43 & 77.79 & 89.12 & 60.16 & 89.44\\
\hline
Precision & 73.67 & 14.80 & 49.69 & 79.29 & 33.19 & 74.9 & 64.19 & 16.38 & 49.86\\
\hline
Recall & 79.49 & 22.37 & 51.55 & 83.35 & 42.40 & 69.93 & 69.50 & 15.88 & 46.18\\
\hline
F1 score & 76.47 & 17.82 & \textbf{50.56} & 81.27 & 37.24 & \textbf{72.33} & 66.74 & 16.13 & \textbf{47.95}\\
\hline
\end{tabular}
\end{table*}

\subsection{Testing different embedding reconstruction loss functions}
\label{subsec:exp4}

To find the best loss function to be used in Eq.~\eqref{eq:minimize_loss} for the activity recognition task, a variety of loss functions were tested on the Opportunity dataset for the unsupervised training of $M_T$. As shown in Table \ref{tab:loss_compare}, MAE loss along with L2 regularization achieved the highest performance metrics among all the functions tested. MAE likely works well because it measures the average error between the source embeddings $e_S$ and the transformed target embeddings $e_T$ without overemphasizing on the point deviations in the two distributions as in some of the loss functions e.g., MSE. 
Furthermore, adding either L1 or L2 regularization shows performance improvement in almost all cases tested. This is expected as regularization helps reduce overfitting on the non-overlapping data between the embeddings $e_S$ and $e_T$. These test results directed our choice of using MAE as the loss function for training $M_T$ for the PAMAP2 and MHEALTH datasets, and for all the other experiments detailed in this paper.

\begin{table*}[!t]
\centering
\caption{Performance metrics when M2 was trained using different loss functions for Opportunity dataset. Acc stands for accuracy, P stands for precision, R stands for recall, F1 stands for F1 score, sim stands for similarity, and reg stands for regularization. Mean absolute error loss along with L2 regularization was seen to provide the best performance.}
\label{tab:loss_compare}
\begin{tabular}%
{|p{0.9cm}|p{0.6cm}|p{0.9cm}|p{1.2cm}|p{0.6cm}|p{0.77cm}|p{0.6cm}|p{0.77cm}|p{0.77cm}|p{0.65cm}|p{0.65cm}|p{0.77cm}|p{0.65cm}|p{0.65cm}|}
\hline
Metrics& $M_S$ & Random samp &Untrained & MSE &MSLE & MAE &Cosine sim &Cosine sim + L1 reg & MAE + L1 reg & MSE + L1 reg &Cosine sim + L2 reg & MAE + L2 reg & MSE + L2 reg\\
\hline
Acc & 76.13 & 26.6 & 24.01 & 52.98 & 54.21 & 49.75 & 59.53 & 60.17 & 64.13 & 62.69 & 57.58 & \textbf{65.13} & 49.10\\
\hline
P & 73.28 & 20.42 & 14.01 & 44.96 & 47.82 & 46.48 & 54.24 & 57.34 & 59.00 & 66.40 & 54.94 & \textbf{66.50} & 36.99\\
\hline
R & 79.60 & 20.48 & 26.2 & 50.03 & 50.83 & 51.51 & 56.36 & 57.16 & 64.17 & 59.82 & 57.24 & \textbf{67.89} & 43.27\\
\hline
F1 & 76.31 & 20.45 & 18.26 & 47.36 & 49.28 & 48.87 & 55.28 & 57.25 & 60.93 & 62.94 & 56.07 & \textbf{67.19} & 39.89\\
\hline
\end{tabular}
\end{table*}

\begin{figure}[t!]
     \centering
     \begin{subfigure}[b]{0.33\textwidth}
         \centering
         \includegraphics[width=\textwidth]{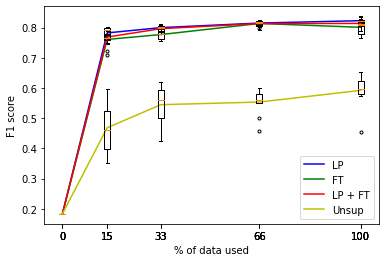}
         \caption{Opporunity dataset}
         \label{fig:opp_5lab_f1_all_methods}
     \end{subfigure}
     \begin{subfigure}[b]{0.33\textwidth}
         \centering
         \includegraphics[width=\textwidth]{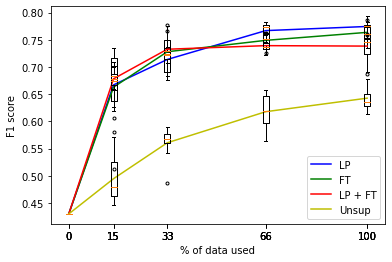}
         \caption{PAMAP2 dataset}
         \label{fig:pamap_5lab_f1_all_methods}
     \end{subfigure}
     \begin{subfigure}[b]{0.33\textwidth}
         \centering
         \includegraphics[width=\textwidth]{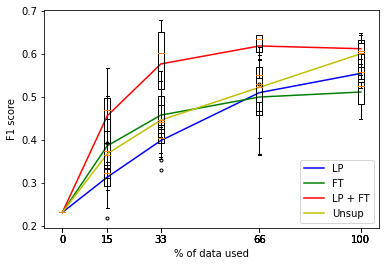}
         \caption{MHEALTH dataset} 
         \label{fig:mhealth_5lab_f1_all_methods}
     \end{subfigure}
        \caption{Comparison of F1 score obtained from proposed unsupervised method and some common transfer learning techniques as detailed in  \cite{kumar2022fine} for different sizes of datasets when five common labels are used. LP stands for linear probing, FT for full fine tuning and LPFT for linear probing with fine tuning. It can be seen that generally transfer learning methods perform better than the proposed unsupervised method. However, this gap decreases as amount of data available decreases and the unsupervised method beats the transfer learning method at very low amounts of data. }
        \label{fig:tf_compare}
\end{figure}

\subsection{Comparison with transfer learning techniques}
\label{subsec:exp5}
The following experiment was conducted to answer the following two questions: 1) when is it worth labeling data in the target domain and using transfer learning techniques over our unsupervised technique of Sect.~\ref{sec:approach}?, and 2) how much labeled data are required for transfer learning techniques to outperform our method? We tested the classification performance of three standard transfer learning techniques: linear probing (LP), full fine tuning (FT) and linear probing along with fine tuning (LPFT) in accordance with Kumar et. al. \cite{kumar2022fine}. LP involves retraining just the classification layer $C_S$ with labeled target domain data. FT entails retraining the entire model $M_S$ using labeled target domain data. In LPFT, the classification layer $C_S$ is first retrained with labeled target domain data and then FT of the entire model is performed. We ran these transfer learning techniques as well as our unsupervised training method using 0\%, 15\%, 33\%, 66\% and 100\% of the training data from each dataset, where 0\% refers to classification without any sort of retraining, i.e. applying $M_S$ on $D_{ST}$. Each of these tests was run 10 times with random sampling of the available training data in order to avoid overfitting. Performance metrics averaged over the 10 runs are reported here. \\

Fig.~\ref{fig:tf_compare} shows F1 scores against the percentage of data used for the three transfer learning techniques and the unsupervised domain transfer approach. It is immediately noticeable that for each approach, the performance on each of the datasets increases or plateaus as the amount of training data increases. All three transfer learning approaches outperformed our unsupervised method for all amounts of data in the Opportunity and PAMAP2 datasets; however, the performance difference between the transfer learning and the unsupervised methods was lower in the PAMAP2 dataset than in the Opportunity dataset. Note that the absolute number of samples in the PAMAP2 dataset is significantly lower than that in the Opportunity dataset. For the MHEALTH dataset, the performance of the unsupervised method is better than LP for all amounts of data, comparable to FT for 15\%, 33\% of the data, and better than FT for 66\% and 100\% of the data. Note that the MHEALTH dataset has the fewest data samples among the three datasets, suggesting that when only limited amount of data are available, there is scope for the unsupervised method to outperform some transfer learning methods. Thus, our method would outperform these transfer learning methods when very little data is available at the target body site, and these transfer learning methods would outperform our method as the amount of labeled data increases. There is a trade-off between performance and the cost of labeling data. The decision of using the unsupervised method or investing in labeling data would depend on the particular use case. 

\subsection{Using all activity classification labels in PAMAP2 and MHEALTH}
\label{subsec:exp6}

Instead of just the five overlapping activity classes, we further tested our method on all the available labels in the PAMAP2 and MHEALTH. 
This was not done for the Opportunity as it only contains the five overlapping labels tested.
As shown in Table \ref{tab:all_label_results}, the performance metrics for $M_S$ on $D_S$ drop for both datasets when compared to those when only 5 labels were used. This is expected because the addition of more classes makes the classification problem more difficult. However, the trend of $M_T$ performing better than $M_S$ on $D_{ST}$ is still visible, suggesting that even for more challenging activity classification tasks, unsupervised training of $M_S$ on $D_{ST}$ helps capture overlapping information in the source and target domain embeddings. Detailed results of comparing our unsupervised method with transfer learning methods when all labels are used can be found in Appendix~\ref{appendix:tf}.

\subsection{Comparison with STranGAN approach}
STranGAN \cite{faridee2022strangan} is a method to perform unsupervised domain transfer between the source and target domain using a spatial transformer. Specifically, the spatial transformer learns affine transformations on the target domain IMU in order to have discriminative features similar to that of the source domain IMU. The spatial transformer is trained together with the domain discriminator which guesses the origin of the transformed IMU signals, i.e. guesses whether the transformed IMU signals belong to the source or target domain. Thus the domain discriminator is adversarial in nature to the spatial transformer. In this experiment, We compared the F1 scores obtained for the Opportunity and PAMAP2 datasets with the results of STranGAN. Note that we used all the activity labels in the PAMAP2 dataset. As shown in Table  \ref{tab:f1_comparison}, our method is comparable to or performs better than STranGAN for the two datasets tested. However, our method provides the key advantage of being able to use any existing activity classifier trained on the source domain, thus carrying over their useful characteristics to the target domain.

\begin{table*}[!t]
\centering
\caption{Performance metrics for PAMAP2 and MHEALTH datasets when all activity labels were used. The performance of $M_S$ on $D_S$ decreases as compared to \ref{tab:5class_results} due to increase in the number of classes. The trend of $M_T$ on $D_{ST}$ performing better than $M_S$ on $D_{ST}$ is seen in this case as well.}
\label{tab:all_label_results}
\begin{tabular}{|p{1.4cm}|p{1.4cm}|p{1.4cm}|p{1.4cm}|p{1.4cm}|p{1.4cm}|p{1.4cm}|}
\hline
Metrics & \multicolumn{3}{|c|}{PAMAP2} & \multicolumn{3}{|c|}{MHEALTH} \\
\hline
  & $M_S$ on $D_S$ & $M_S$ on $D_{ST}$ & $M_T$ on $D_{ST}$ & $M_S$ on $D_S$ & $M_S$ on $D_{ST}$ & $M_T$ on $D_{ST}$\\
\hline
Accuracy & 64.66 & 33.55 & 58.50 & 76.99 & 58.35 & 79.05\\
\hline
Precision & 66.57 & 16.54 & 60.96 & 57.39 & 18.68 & 57.03\\
\hline
Recall & 73.66 & 21.87 & 61.78 & 87.09 & 32.12 & 78.16\\
\hline
F1 score & 69.94 & 18.84 & \textbf{61.37} & 69.19 & 23.63 & \textbf{65.95}\\
\hline
\end{tabular}
\end{table*}

\begin{table}[!t]
\centering
\caption{Comparison of F1 scores for our method with the results reported for the Opportunity and PAMAP2 datasets in \cite{faridee2022strangan} using the STranGAN method. Our method is comparable to or performs better than \cite{faridee2022strangan}.}
\label{tab:f1_comparison}
\begin{tabular}{|p{1.6cm}|p{1.6cm}|p{1.6cm}|p{1.6cm}|p{1.6cm}|}
\hline
 & \multicolumn{2}{|c|}{Opportunity} & \multicolumn{2}{|c|}{PAMAP2} \\
\hline
  & STranGAN & Ours & STranGAN & Ours\\
\hline
F1 score & 62.6 & \textbf{67.19} & \textbf{64.86} & 61.37\\
\hline
\end{tabular}
\end{table}

\section{Conclusion}
As activity recognition using wearable devices is gaining more applications in everyday scenarios, the flexibility of using such devices on different body locations is becoming important. In this work, we proposed an unsupervised approach to leverage an existing activity classification model, which is trained on a source body location, to perform activity classification at a target body location. We have shown that a pre-trained activity classifier can be used to perform activity classification at a new target body location if the classifier is retrained to replicate the source body location embeddings using target body location signals. The performance of the retrained classifier depends on the performance of the original classifier on the source body location. The performance of the retrained model also increases as the amount of simultaneous source-target domain data increases. Hence, the need of labeled data at the target body location is removed, significantly saving time and cost because collecting simultaneous measurements at the source and target body locations is far less expensive. Additionally, leveraging an existing activity classifier that has proven to work well on one body location allows us to retain important properties of the model, such as the model being lightweight, which is critical for running the model on edge devices like smartwatches.

\bibliographystyle{unsrt}  
\bibliography{arxiv_article}  

\section*{Appendix}
\renewcommand{\thesubsection}{\Alph{subsection}}

\subsection{Confusion matrices} \label{appendix:confusion}
\begin{figure*}
    \centering
     \begin{subfigure}[tb]{0.45\textwidth}
         \centering
         \includegraphics[width=0.9\textwidth]{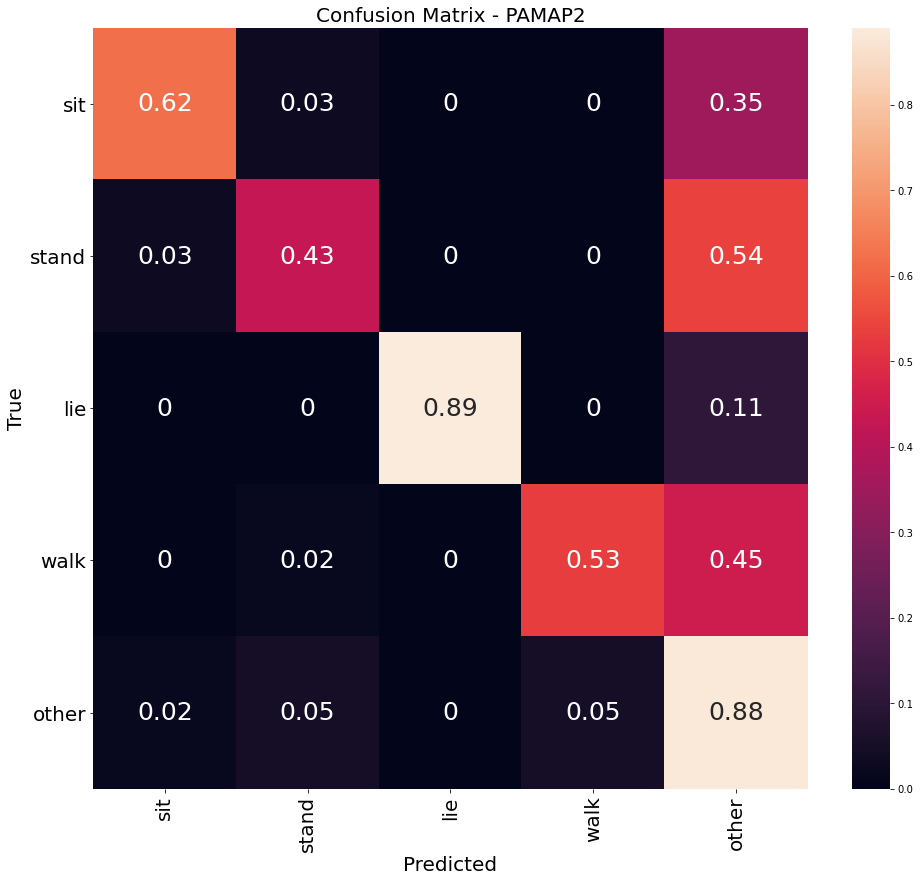}
         \caption{PAMAP2 dataset}
         \label{fig:confusion_pamap}
     \end{subfigure}
     \begin{subfigure}[tb]{0.45\textwidth}
         \centering
         \includegraphics[width=0.9\textwidth]{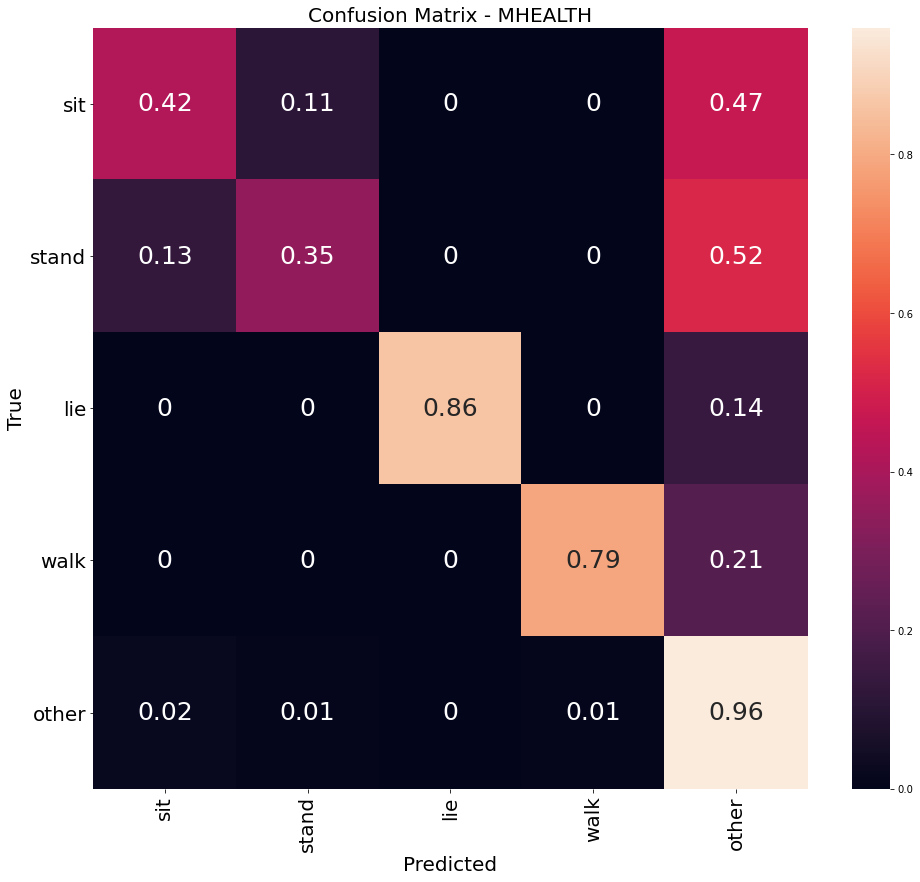}
         \caption{MHEALTH dataset} 
         \label{fig:confusion_mhealth}
     \end{subfigure}
    \caption{Confusion matrices for PAMAP2 and MHEALTH datasets.}
    \label{fig:confusion}
\end{figure*}

The confusion matrices for PAMAP2 and MHEALTH datasets are shown in Fig. \ref{fig:confusion}. It can be seen that for both datasets,  \emph{other} class is classified correctly to the most extent but all the remaining classes are confused for \emph{other} often. This might be due to the high number of \emph{other} samples in the datasets.

\subsection{ROC curves} \label{appendix:roc}
\begin{figure*}[!t]
    \centering
     \begin{subfigure}[!t]{0.45\textwidth}
         \centering
         \includegraphics[width=\textwidth]{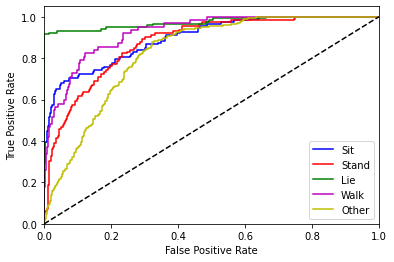}
         \caption{PAMAP2 dataset}
         \label{fig:roc_pamap}
     \end{subfigure}
     \begin{subfigure}[!t]{0.45\textwidth}
         \centering
         \includegraphics[width=\textwidth]{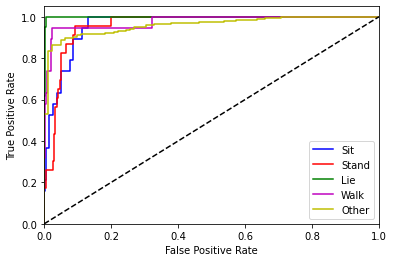}
         \caption{MHEALTH dataset} 
         \label{fig:roc_mhealth}
     \end{subfigure}
    \caption{ROC curves for each label for all three datasets when five labels are used.}
    \label{fig:roc}
\end{figure*}

The ROC curves for PAMAP2 and MHEALTH are shown in Fig. \ref{fig:roc}. It can be seen that the ROC curves for the MHEALTH dataset, especially for the \emph{lie} class, are near perfect. However, it is important to remember that the dataset, and test set specifically, had very few samples of each class other than \emph{other}. In both datasets, \emph{lie} class has the highest AUC and low AUC for \emph{other} as reflected in \ref{appendix:confusion}. 

\begin{figure*}[!t]
    \includegraphics[width=0.95\textwidth]{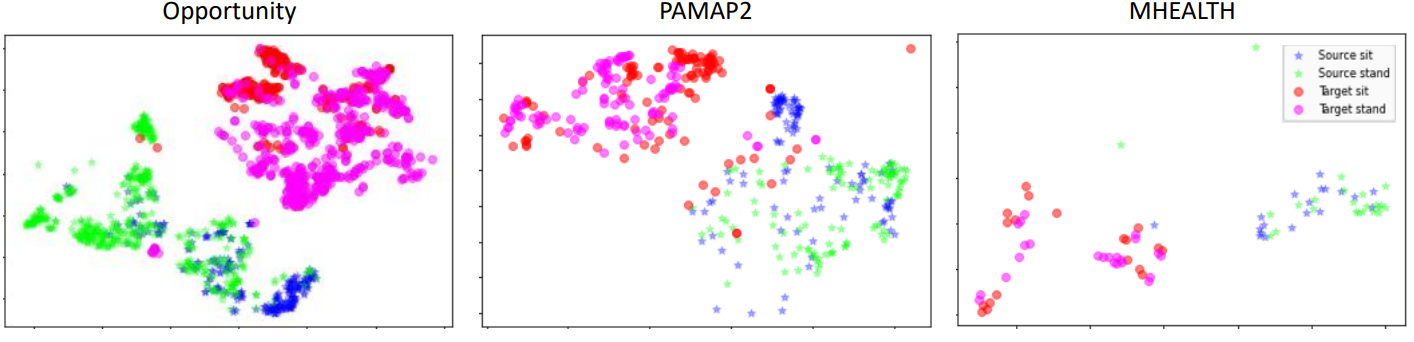}
    \centering
    \caption{t-SNE projections for the labels \emph{sit} and \emph{stand} for source and target domains for all three dataset after training of $M_T$. The source and target projections are fairly clearly differentiated. The separation between \emph{sit} and \emph{stand} is a little less clear but is still visible.} 
    \label{fig:tsne}
\end{figure*}

\begin{figure*}[t!]
     \centering
     \begin{subfigure}[t]{0.45\textwidth}
         \centering
         \includegraphics[width=\textwidth]{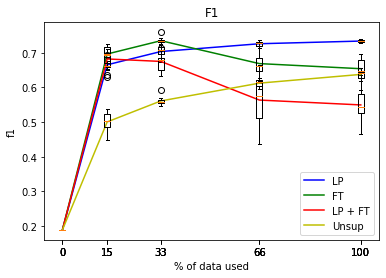}
         \caption{PAMAP2 dataset} 
         \label{fig:pamap_all_labels_f1_all_methods}
     \end{subfigure}
     \begin{subfigure}[t]{0.45\textwidth}
         \centering
         \includegraphics[width=\textwidth]{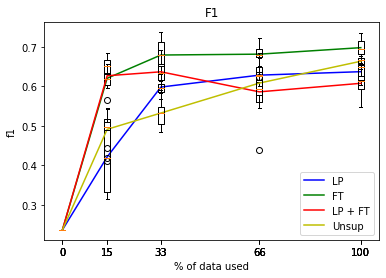}
         \caption{MHEALTH dataset}
         \label{fig:mhealth_all_labels_f1_all_methods}
     \end{subfigure}
     \caption{Comparison of F1 score obtained from proposed unsupervised method and some common transfer learning techniques for different sizes of datasets when all labels are used for PAMAP2 and MHEALTH datasets. LP stands for linear probing, FT for full fine tuning and LPFT for linear probing with fine tuning. It can be seen that generally transfer learning methods perform better than the proposed unsupervised method. However, this gap decreases as amount of data available decreases and the unsupervised method beats the transfer learning method at very low amounts of data.}
     \label{fig:all_labels_tf_compare}
\end{figure*}
\subsection{t-SNE projections} \label{appendix:tsne}
The t-SNE projections for two labels, \emph{sit} and \emph{stand}, for source and target domains are shown in Fig. \ref{fig:tsne} for each dataset after $M_T$ was trained. It can be seen that source and target projections are fairly clearly differentiated in all datasets. However, the separation between \emph{sit} and \emph{stand} is a little less clear but is still visible. It is also visible that the number of samples for MHEALTH dataset are low compared to the other two datasets.

\subsection{Comparison with transfer learning methods when all labels used} \label{appendix:tf}
We compared the performance of the unsupervised method with transfer learning techniques LP, FT and LPFT as detailed in \cite{kumar2022fine} when all the labels are used for PAMAP2 and MHEALTH datasets. This is shown in Fig. \ref{fig:all_labels_tf_compare}. It can be seen that generally transfer learning methods perform better than the proposed unsupervised method. However, this gap decreases as amount of data available decreases and the unsupervised method beats the transfer learning method at very low amounts of data.

\end{document}